\pdfoutput=1

\documentclass[11pt]{article}

\usepackage[]{acl}

\usepackage{times}
\usepackage{latexsym}

\usepackage{multirow}
\usepackage{booktabs}
\usepackage[T1]{fontenc}

\usepackage[utf8]{inputenc}

\usepackage{microtype}

\usepackage{inconsolata}

\usepackage{graphicx}

\title{Break the Chain: Large Language Models Can be Shortcut Reasoners}

\author{Mengru Ding $^*$\\
  Gaode Map, Alibaba Group  \\
  \texttt{dingmengru.dmr@alibaba-inc.com} \\
  \And
  Hanmeng Liu $^*$\\
  Zhejiang University  \\
  \texttt{liuhanmeng@zju.edu.cn} \\
  \And
  Zhizhang Fu \\
  Gaode Map, Alibaba Group  \\
  \texttt{fuzhizhang.fzz@alibaba-inc.com} \\
  \AND
  Jian Song \and Wenbo Xie  \\
  Gaode Map, Alibaba Group  \\
  \texttt{{james.songjian, daoqian.xwb}@alibaba-inc.com} \\
  \And
  Yue Zhang † \\
  Westlake University  \\
  \texttt{zhangyue@westlake.edu.cn}
  }

\begin{document}
\maketitle

\def\thefootnote{*}\footnotetext{These authors contributed equally to this work}\def\thefootnote{\arabic{footnote}}

\def\thefootnote{†}\footnotetext{Corresponding author}\def\thefootnote{\arabic{footnote}}

\begin{abstract}
Recent advancements in Chain-of-Thought (CoT) reasoning utilize complex modules but are hampered by high token consumption, limited applicability, and challenges in reproducibility. This paper conducts a critical evaluation of CoT prompting, extending beyond arithmetic to include complex logical and commonsense reasoning tasks, areas where standard CoT methods fall short. We propose the integration of human-like heuristics and shortcuts into language models (LMs) through "break the chain" strategies. These strategies disrupt traditional CoT processes using controlled variables to assess their efficacy.  Additionally, we develop innovative zero-shot prompting strategies that encourage the use of shortcuts, enabling LMs to quickly exploit reasoning clues and bypass detailed procedural steps. Our comprehensive experiments across various LMs, both commercial and open-source, reveal that LMs maintain effective performance with "break the chain" strategies.
We also introduce ShortcutQA, a dataset specifically designed to evaluate reasoning through shortcuts, compiled from competitive tests optimized for heuristic reasoning tasks such as forward/backward reasoning and simplification. Our analysis confirms that ShortcutQA not only poses a robust challenge to LMs but also serves as an essential benchmark for enhancing reasoning efficiency in AI.
\end{abstract}

\section{Introduction}

\begin{figure}[h]
    \centering
    \includegraphics[width=0.5\textwidth]{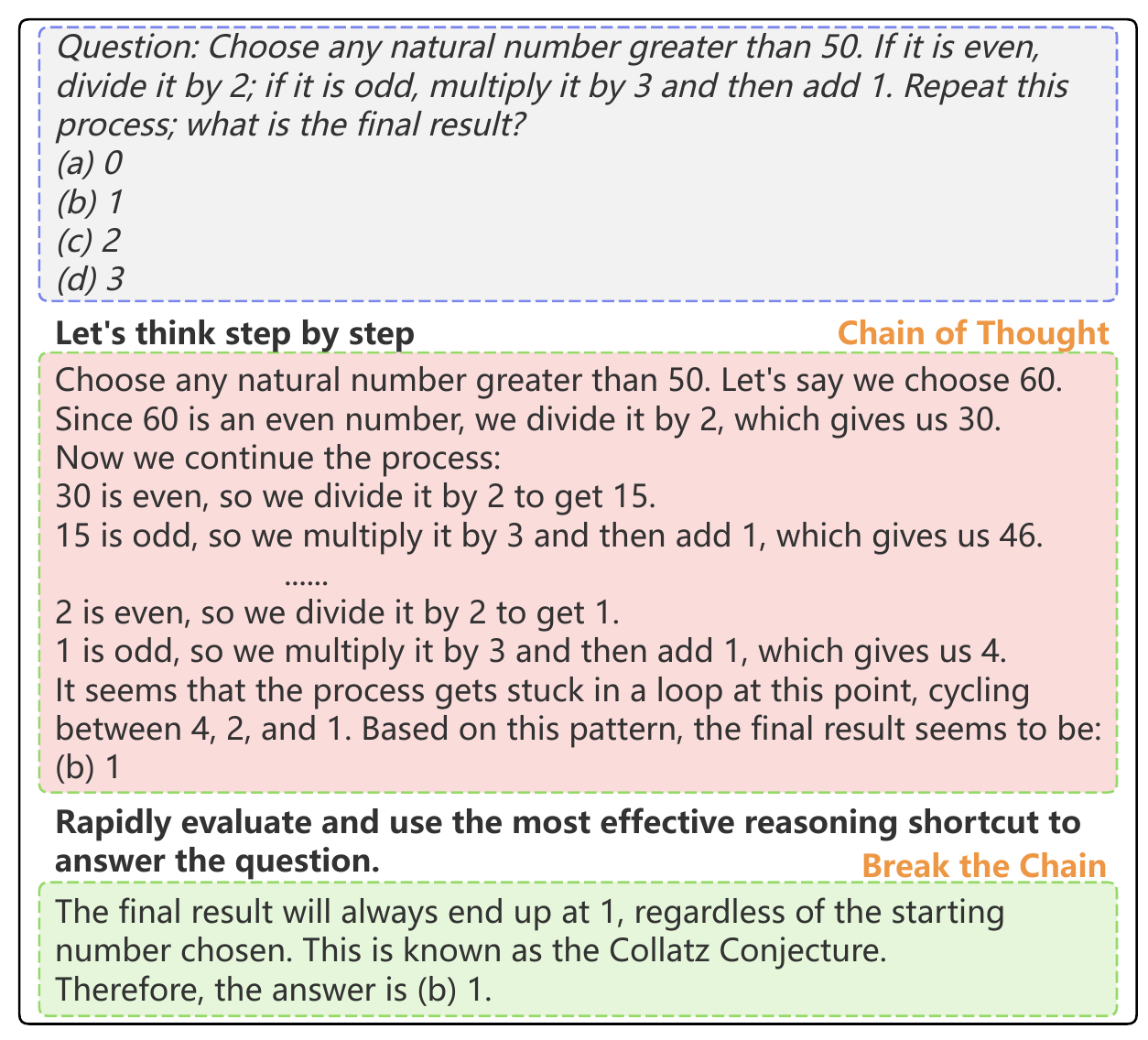}
    \caption{ChatGPT responses to Chain-of-Thought and "Break the Chain". Our "Break the Chain" method significantly simplifies the reasoning process.}
    \label{fig:showcase}
\end{figure}

In the evolving landscape of artificial intelligence, the ability to reason and solve complex problems symbolizes a cornerstone of intelligence. Language Models (LMs), particularly those based on transformer \cite{vaswani2017attention} architectures, have revolutionized our approach to natural language processing (NLP), significantly enhancing capabilities in comprehending and generating text that bears a striking resemblance to human communication.


Among recent advancements, Chain-of-Thought (CoT) prompting has emerged as a pivotal technique for utilizing Large Language Models (LLMs) to address complex reasoning tasks. By methodically eliciting step-by-step reasoning, CoT prompting has significantly enhanced the problem-solving capabilities of LLMs across a variety of learning scenarios, including few-shot \cite{wei2022chain} and zero-shot contexts \cite{kojima2022large}. Figure~\ref{fig:showcase} illustrates a zero-shot example in which the ChatGPT model methodically resolves a mathematical question. This strategy is further augmented by approaches such as self-consistency \cite{wang2022self,wang2023scott}, interactive reasoning \cite{yao2022react,shinn2024reflexion}, reflective thinking \cite{ling2024deductive,li2023reflection}, task decomposition \cite{khot2022decomposed,press2022measuring}, and strategic planning \cite{wang2023plan,hu2023chain}.

Despite its benefits, CoT is also critiqued for its substantial token usage, as it explores numerous reasoning pathways before arriving at a conclusive answer. This characteristic is particularly prominent in variants such as Tree-of-Thought (ToT) \cite{Yao2023TreeOT}, which scrutinize every possible reasoning chain. Traditionally, CoT has been predominantly applied to mathematical reasoning, with scant application to commonsense, or complex logical reasoning tasks. This limited focus may hinder a comprehensive understanding of CoT's potential to emulate intricate human-like reasoning processes. Additionally, instruction fine-tuned (IFT) \cite{ouyang2022training} large language models like ChatGPT, which are usually capable of reaching the answers methodically, further question the necessity for explicit CoT prompting \cite{chen2023you}.


Human reasoning uses heuristics to find local rational maximum \cite{karlan2021reasoning,neth2015heuristics,lancia2023humans},
which often relies on cognitive shortcuts \cite{fernbach2013cognitive,ferrario2004developing}, a characteristic that can be mirrored and exploited in LMs. Traditionally, LLMs' shortcut learning has been viewed as the acquisition of spurious correlations within datasets \cite{10.1145/3596490,jiang2019avoiding,branco2021shortcutted}. However, this perspective fails to capture the nuanced heuristic reasoning processes inherent in human cognition, both in everyday scenarios and professional contexts such as clinical decision-making. We argue that shortcut reasoning, by drastically reducing reasoning steps and computational demands, offers a valuable means of enhancing LLM efficiency. As depicted in Figure~\ref{fig:showcase}, when prompted with shortcut reasoning, the ChatGPT model swiftly arrives at answers with minimal token consumption. The ability of LLMs to employ shortcut reasoning not only mirrors human cognitive strategies but also has the potential to streamline problem-solving processes, thereby reshaping computational efficiency and model performance.

The primary goal of our study is to critically evaluate and challenge the established Chain-of-Thought (CoT) prompting framework used in Large Language Models (LLMs). Our approach is three-pronged: First, we explore the effectiveness, limitations, and mechanisms of CoT by comparing it with different prompts derived from the "break the chain" strategy in both few-shot and zero-shot scenarios. Second, the study pioneers the use of shortcut reasoning prompts that encourage LLMs to utilize heuristic shortcuts — akin to intuitive leaps in human reasoning — to efficiently solve problems. This method aims to minimize computational demands and token consumption while maintaining or potentially enhancing performance accuracy. To support this investigation, we introduce ShortcutQA, a novel dataset meticulously curated to specifically assess the ability of LLMs to employ heuristic shortcuts. We conducted experiments on both OpenAI models and open-source models of various sizes, including \textsc{Mixtral-8x7B-Instruction}, \textsc{Llama-3-70B-Instruction}, \textsc{Qwen1.5-72B-Chat}, \textsc{Qwen1.5-14B-Chat}, \textsc{Qwen1.5-1.8B-Chat}, to ascertain the generalizability of our experimental conclusions across different model configurations.

Our few-shot experiments reveal that Large Language Models (LLMs) are not adversely affected by disrupted Chain-of-Thought (CoT) demonstrations, casting doubts on the effectiveness of few-shot CoT methods. To our knowledge, this is the first series of experiments designed to "break the chain" of in-context examples. Furthermore, in zero-shot scenarios, models prompted with shortcut reasoning display robust performance, often surpassing that of traditional CoT methods. Our evaluations span both OpenAI models and open-source models, showing consistent results across platforms.

Furthermore, our comparative analysis elucidates distinct performance trends across various model sizes: smaller models typically experience more substantial enhancements with Chain-of-Thought (CoT) prompts compared to their larger counterparts. Notably, as model size increases, the efficacy of "break the chain" strategies becomes more pronounced, highlighting its effectiveness in mitigating the impact of disrupted CoT demonstrations. 

Most notably, we observe that shortcut reasoning significantly reduces token consumption, providing a vital advantage in computational efficiency. Under stringent token constraints, shortcut reasoning strategies not only conserve resources but also consistently outperform traditional CoT methods. These benefits are observed across various datasets, underscoring the robustness and scalability of shortcut reasoning as a superior approach in enhancing LLM performance.

\section{Related Work}

\subsection{CoT Prompting in Large Language Models}
The evolution of Chain-of-Thought (CoT) prompting, particularly through few-shot~\cite{wei2022chain} and zero-shot~\cite{kojima2022large} methodologies, has markedly advanced Large Language Models' (LLMs) ability to address complex reasoning challenges. This field has witnessed the introduction of sophisticated data structures, such as Tree-of-Thought~\cite{Yao2023TreeOT}, Graph-of-Thought~\cite{besta2024graph}, and Program-of-Thought~\cite{chen2022program}, enriching LLMs' capacity for introspection and nuanced evaluation of their reasoning paths.

Beyond conventional prompting strategies, the ReAct model~\cite{Yao2022ReActSR} integrates reasoning with actionable tasks like data retrieval, whereas the Selection-Inference framework~\cite{creswell2023selectioninference} combines context creation with logical chaining. While pioneering, these approaches rely on the models' inherent abilities and do not embed explicit logical rules within the reasoning process.

The adoption of external tools in prompting paradigms, especially for tasks that demand supplementary knowledge, has also shown considerable progress. Analogous to the role calculators play in mathematical reasoning, introducing predefined functions for enforcing inference rules marks a significant step forward in leveraging external computational aids to bolster reasoning capabilities.

Moreover, breaking down complex reasoning tasks into more manageable subproblems or engaging multiple models for collaborative problem-solving has introduced novel methodologies in LLM prompting. Strategies such as Cumulative Reasoning~\cite{Zhang2023CumulativeRW} focus on an iterative, step-wise approach, while ScratchPad~\cite{nye2021show} emphasizes the articulation of intermediate steps in multi-step reasoning. Meta-prompting~\cite{suzgun2024meta} envisions a cooperative framework where LLMs act as orchestrators, decomposing tasks, delegating them to specialized models, and synthesizing the outcomes, thereby fostering a holistic approach to problem-solving.

In the specific arena of instruct-tuning LLMs with tailored datasets for advanced reasoning, initiatives like LogiCoT~\cite{liu2023logicot}, which fine-tunes an LLaMA-7b model with data on logical chaining, demonstrate considerable improvements in logical reasoning tasks. Similarly, LogicLLM~\cite{jiao2023logicllm} explores a self-supervised learning strategy for logical reasoning enhancements, and Symbol-LLM~\cite{xu2023symbolllm} incorporates symbolic data in a two-stage tuning process to equip a LLaMA-2-chat model with symbolic reasoning skills. These efforts highlight the potential of fine-tuning with specialized datasets to significantly enhance the reasoning capabilities of LLMs, illustrating the dynamic and evolving landscape of CoT prompting in AI research.

\subsection{Questioning CoT}
Despite the demonstrated effectiveness of Chain-of-Thought (CoT) in enhancing model performance on complex tasks, the underlying mechanisms by which Large Language Models (LLMs) generate CoT responses are not fully understood. Research efforts are increasingly focused on demystifying CoT prompting, providing empirical insights and developing theoretical frameworks to comprehend this advanced reasoning capability. However, numerous studies have highlighted the brittleness of CoT reasoning in various aspects.

\citet{NEURIPS2023_ed3fea90} investigate the faithfulness of CoT reasoning, revealing systematic misrepresentations in the true rationale behind a model’s predictions. \citet{lanham2023measuring} extend this inquiry by introducing errors or paraphrases within the CoT process to test whether the articulated reasoning truly reflects the model’s underlying logic, finding that larger models tend to produce more unfaithful responses. This issue of faithfulness is critical as it challenges the reliability of CoT explanations.
The effectiveness of CoT is also impacted by the selection and arrangement of demonstrations. \citet{CoTEmpiricalStudy:BoshiWang} find that the accuracy of reasoning chains is less critical than the relevance of the question and the correctness of the reasoning sequence, emphasizing the importance of contextual alignment. In contrast, \citet{wang2022towards} show that CoT can operate even with invalid demonstrations, suggesting some resilience in the reasoning process. Our research contributes to this discourse by disturbing the order of the reasoning chain to examine its impact on CoT consistency.

\begin{table*}
\centering
\scalebox{0.8}{
\begin{tabular}{c|c|c|c|c}
 \toprule
 Dataset  & Question Type  & \# of instances & Avg. \# words & Source \\ 
 \midrule
 \multirow{7}{*}
             & Analytical shortcuts      & 156 & 55.88 & Analytical reasoning tests \\ 
  ShortcutQA & Logical shortcuts      & 108 & 21.76  & Verbal reasoning tests \\ 
             & Mathematical shortcuts  & 185 & 67.19  & Gaokao examinations \\ 
 \bottomrule
\end{tabular}
}
\caption{Dataset statistics of ShortcutQA. }
\label{table:shortcutQA}
\end{table*}

\citet{jin2024impact} demonstrate that artificially lengthening the reasoning steps in prompts — simply by instructing models to "think more steps" — can enhance LLMs' performance across various datasets without introducing new content. This finding suggests that the perceived depth of reasoning may artificially inflate effectiveness. Conversely, we explore minimalist prompting strategies where LLMs are instructed to streamline their reasoning processes.

The sensitivity of LLMs to the ordering of premises is scrutinized by \citet{chen2024premise}, who note optimal performance when the order of premises supports the necessary context in intermediate reasoning steps. This sensitivity is paradoxical in deductive reasoning contexts where the order of premises should not logically influence the validity of conclusions. Similarly, \citet{pfau2024lets} Indicates that LLMs solve more problems with meaningless filler tokens in place of a chain of thought than without meaningless tokens. This finding suggests that CoT's effectiveness may sometimes rely solely on the increase in computational effort, rather than on the literal intermediate reasoning steps. Our "break the chain" methods experiment with new models and datasets and aim to illuminate this issue further.

Implicit CoT \cite{deng2023implicit,deng2024explicit} has been introduced to internalize explicit step-by-step reasoning. Similar to our work, implicit CoT questions the necessity of step-by-step reasoning. However, we diverge from prior studies that employed fine-tuning to reduce the need for reasoning steps.

Finally, \citet{chen2023you} question the applicability of CoT in instruction fine-tuned (IFT) models like ChatGPT, which show inconsistent performance across various reasoning tasks. Surprisingly, while CoT prompts enhance some reasoning tasks, they fail in others like arithmetic reasoning, where ChatGPT can independently generate CoT sequences without specific prompts. This phenomenon inspires us to abstract a hypothesis that more powerful models increasingly exhibit a reduced dependency on CoT. Our subsequent experiments conducted within the Qwen1.5 series of various sizes strive to support this viewpoint.


\section{ShortcutQA}

The ShortcutQA dataset is designed to evaluate Language Models' (LMs) ability to employ heuristic shortcuts in reasoning, addressing a gap in existing resources that primarily focus on sequential reasoning approaches. Comprising 449 diverse reasoning problems, ShortcutQA spans logical puzzles to real-world problem-solving scenarios. Each problem is presented with a shortcut-based solution alongside a detailed step-by-step solution, categorized into three reasoning types.

\textbf{Data Collection and Annotation}

Data for ShortcutQA were sourced from various online forums and educational websites, with necessary permissions secured. Annotation was conducted by two independent domain experts, adhering to strict guidelines for identifying and categorizing heuristic shortcuts employed in the solutions. A third expert resolved any discrepancies, ensuring high annotation quality and consistency.

\textbf{Dataset Categorization}

ShortcutQA introduces problems categorized into three distinct types, each testing different aspects of heuristic reasoning:
\begin{itemize}
    \item \textbf{Analytical Shortcuts:} Tasks necessitate analyzing situations beyond mere comprehension, assessing models' capabilities in efficiently synthesizing and utilizing key information, and strategic decision-making under time constraints.
    \item \textbf{Logical Shortcuts:} Encompassing forms of reasoning such as analogical, abductive, and forward/backward reasoning, these tasks focus on applying these logical theories to derive conclusions from provided statements.
    \item \textbf{Mathematical Shortcuts:} Features problems solvable through approximation techniques, substitution, simplification, and special-case reasoning, bypassing traditional sequential thought processes.
\end{itemize}

Data Statistics are shown in Table~\ref{table:shortcutQA}. We release the data at https://anonymous.com.

\section{Method}
\subsection{Break the Chain}
To examine the resilience and limitations of Large Language Models (LLMs) in employing Chain-of-Thought (CoT) reasoning, our research outlines a novel experimental framework aimed at "breaking the chain" of thought. This approach seeks to elucidate the conditions under which CoT reasoning may falter, thereby offering insights into the underlying mechanisms of LLMs' reasoning capabilities. Our methodology juxtaposes zero-shot and few-shot scenarios to delineate the impact of CoT disruption across different prompting contexts.

\textbf{Few-Shot}
In the few-shot scenario, our strategy involves perturbing the sequence of sentences within the in-context examples provided to the LLM. This disturbance is designed to misalign the logical progression typically demonstrated in CoT reasoning, thereby testing the model's ability to maintain coherent and accurate reasoning despite the disordered presentation of steps. This manipulation will help ascertain the significance of stepwise logical progression in the model's reasoning efficacy and its ability to reorient itself to reach correct conclusions.

\textbf{Zero-Shot}
We initiate probing experiments to assess the efficacy of zero-shot CoT prompts, aiming to discern whether CoT prompting is essential or merely a byproduct of longer model responses. Employing controlled experiments, we craft prompts that obviate the need for reasoning chains, instructing models to provide either more verbose or minimalist responses. Detailed descriptions of these prompts are provided in Appendix~\ref{app:prompts}.
Furthermore, we employ meticulously designed prompts to stimulate shortcut reasoning, outlined comprehensively in Appendix~\ref{app:prompts}. By directing LLMs to circumvent intermediate reasoning steps typically associated with CoT, we aim to evaluate the resilience of their inferential processes and their reliance on detailed reasoning pathways.

\textbf{ShortcutQA Probing}
Parallel to our few-shot and zero-shot experiments, we introduce the ShortcutQA dataset into our methodology. ShortcutQA is carefully curated to focus on questions that require shortcut reasoning — a form of intuitive problem-solving that deviates from traditional step-by-step logical deduction. The inclusion of ShortcutQA is intended to test the hypothesis that LLMs can effectively employ heuristic shortcuts, akin to human cognitive shortcuts, to efficiently resolve complex problems.

\subsection{Experimental Setup}
We evaluate Large Language Models (LLMs) across a variety of commercial and open-source platforms under both few-shot and zero-shot conditions. Our methodology includes a diverse array of complex problem-solving tasks encompassing arithmetic reasoning, commonsense deduction, and logical reasoning. This design rigorously tests the LLMs' ability to generalize across different difficulty levels and domains.

\begin{center}
\begin{table}[h!]
\scalebox{0.75}{
\begin{tabular}{ l | c | c | c } 
 \hline
 \textbf{Task} & \textbf{Dataset}  & \textbf{Size} & \textbf{Avg \#words}  \\ 
 \hline
 \multirow{6}{*}{Arithmetic} & SingleEq  & 508 & 27.4 \\ 
  & AddSub  & 395 & 31.5 \\ 
  & MultiArith  & 600 & 31.8 \\
  & GSM8K  & 1319 & 46.9 \\
  & AQUA-RAT  & 254 & 51.9 \\
  & SVAMP  & 1000 & 31.8 \\
 \hline
 \multirow{2}{*}{Commonsense} & CommonsenseQA  & 1221 & 27.8 \\
  & StrategyQA  & 2290 & 9.6 \\
 \hline
 \multirow{4}{*}{Logic} & Date Understanding  & 369 & 35.0 \\
  & Coin Flip  & 500 & 37.0 \\
  & LogiQA  & 651 & 146.2 \\
  & ReClor  & 500 & 153.0\\
 \hline
\end{tabular}
}
\caption{Statistics of Evaluation benchmarks. 
}
\label{table:2}
\end{table}
\end{center}

As depicted in Figure~\ref{fig:method} in Appendix~\ref{app:pipeline}, the experimental pipeline begins by inputting a question and a prompt into an LLM, which then generates a reasoned response and answer. This output is concatenated with the original question and prompt, followed by an answer extraction prompt to extract the final answer.

\begin{table*}
\centering
\scalebox{0.8}{
\begin{tabular}{c|c|c|c|c|c|c}
 \toprule
 \multirow{2}{*}{Task} & \multirow{2}{*}{Dataset} & \multicolumn{2}{c|}{Few-shot} & \multicolumn{3}{c}{Zero-shot} \\ \cline{3-7} 
                       &                          & Base                      & Break the Chain       & Base & No Steps & More Tokens           \\
 \midrule
 \multirow{6}{*}{Arithmetic} & SingleEq & \textbf{92.72} & 92.32 & 86.61 & \textbf{90.35} & 88.39 \\ 
  & AddSub & 84.05 & \textbf{85.32} & 83.80 & \textbf{89.62} & 86.58 \\ 
  & MultiArith & \textbf{99.00} & 98.33 & 83.33 & 91.17 & \textbf{93.50} \\
  & GSM8K & \textbf{74.60} & 74.22 & 32.68 & 37.53 & \textbf{38.89} \\
  & AQUA-RAT & 53.15 & \textbf{55.51} & 35.43 & 36.61 & \textbf{38.97} \\
  & SVAMP & 76.80 & \textbf{79.70} & 71.70 & \textbf{81.70} & 76.70 \\
 \midrule
 \multirow{2}{*}{Commonsense} & CommonsenseQA & 74.94 & \textbf{75.18} & 70.52 & \textbf{75.92} & 74.28 \\
  & StrategyQA & \textbf{69.13} & 68.60 & 64.37 & 59.91 & \textbf{63.23} \\
 \midrule
 \multirow{3}{*}{Logic} & Date Understanding & 81.03 & \textbf{82.11} & 64.37 & \textbf{64.50} & 63.23 \\
  & LogiQA & \textbf{35.94} & 33.95 & 40.09 & \textbf{41.17} & 40.09 \\
  & ReClor & \textbf{51.40} & 50.80 & 52.40 & 51.20 & \textbf{54.20} \\
 \bottomrule
\end{tabular}
}
\caption{ChatGPT performance comparison across tasks. All results are in \%, the best ones are in \textbf{bold}.}
\label{table:btc}
\end{table*}

\paragraph{Benchmarks}
For \textit{arithmetic reasoning}, we assess the models using six datasets: SingleEq~\cite{SingleEQ:Koncel-Kedziorski}, AddSub~\cite{AddSub:MohanmmadJavad}
, MultiArith~\cite{MultiArith:RoySubhro}
, GSM8K~\cite{Verifier:KarlCobbe}, AQUA-RAT~\cite{AQuA:WangLing}, and SVAMP~\cite{patel2021nlp}. The first three originate from the well-established Math World Problem Repository~\cite{koncel-kedziorski-etal-2016-mawps}, with the remaining datasets presenting more recent and complex challenges. SingleEq and AddSub feature relatively straightforward problems that can be solved without multi-step reasoning, whereas MultiArith, AQUA-RAT, GSM8K, and SVAMP require more intricate, sequential problem-solving.

For \textit{commonsense reasoning}, we utilize the CommonsenseQA~\cite{CommonsenseQA:AlonTalmor} and StrategyQA~\cite{geva2021did} datasets. CommonsenseQA tests reasoning based on general world knowledge~\cite{CommonsenseQA:AlonTalmor}
, while StrategyQA demands inference of unstated, multi-step reasoning processes~\cite{geva2021did}.

For \textit{logical reasoning tasks}, we select two scenarios from the BIG-bench~\cite{srivastava2022beyond}: Date Understanding and Coin Flip~\cite{wei2022chain}. Date Understanding challenges models to infer dates from given contexts, and Coin Flip evaluates the ability to determine the outcome of a series of coin flips. Additionally, we incorporate LogiQA~\cite{LogiQA:JianLiu} and ReClor~\cite{RClor:WeihaoYu}, which are reading comprehension tests that require logical deduction.

\paragraph{Language Models}
We test both OpenAI commercial models and huggingface open-source models. For OpenAI models, we choose the ChatGPT (gpt-3.5-turbo-0613) model, an IFT GPT-3 model.
For community models, we use Llama-3-70B-Instruct, Mixtral-8x7B-Instruct, Qwen1.5-72B-Chat, Qwen1.5-14B-Chat, Qwen1.5-1.8B-Chat.

\paragraph{Baselines}
We run zero-shot CoT~\cite{ZeroshotCoT:TakeshiKojima} and few-shot CoT~\cite{wei2022chain} on the datasets to establish our baselines. 
In the few-shot CoT setup, we follow~\citet{wei2022chain} to provide each test with context examples;
for the zero-shot baseline, each question is suffixed with ``The answer is '', following prior work~\cite{ZeroshotCoT:TakeshiKojima,AutoCoT:ZhuoshengZhang}. 


\section{Results}

\begin{table*}
\centering
\scalebox{0.8}{
\begin{tabular}{c|c|c|c|c|c|c}
 \toprule
 Model  & Task & Base & Quick Conclude & Shortcut Reasoning & Effective Shortcut & Innovative Shortcut \\ 
 \midrule
 \multirow{3}{*}{ChatGPT} & Arithmetic & 65.59 & 77.23 & 80.11 & \textbf{80.58} & 71.34 \\
                      & Commensense & 67.45 & 73.18 & \textbf{73.65} & 72.36 & 67.52 \\
                    & Logical & 51.97 & 53.32 & \textbf{57.57} & 56.77 & 56.91 \\
 \midrule
 \multirow{3}{*}{Llama-70B} & Arithmetic & 72.47 & 62.29 & \textbf{81.59} & 63.37 & 50.96 \\
                    & Commensense & 67.57 & \textbf{73.00} & 60.58 & 67.29 & 67.27 \\
                    & Logical & \textbf{71.41} & 66.26 & 68.60 & 67.18 & 63.95 \\
 \midrule
 \multirow{3}{*}{Mixtral-8x7B} & Arithmetic & 70.80 & \textbf{73.22} & 71.70 & 68.77 & 56.63 \\
                    & Commensense & 65.03 & 69.37 & 69.23 & \textbf{69.50} & 60.81 \\
                    & Logical & 69.08 & 69.61 & \textbf{69.84} & 68.48 & 60.46 \\
 \midrule
 \multirow{3}{*}{Qwen1.5-72B} & Arithmetic & 65.28 & \textbf{76.00} & 75.51 & 74.52 & 70.83 \\
                    & Commensense & 79.11 & 79.85 & 79.38 & \textbf{80.36} & 79.78 \\
                    & Logical & 60.17 & 63.42 & 63.58 & \textbf{63.62} & 61.79 \\
 \midrule
 \multirow{3}{*}{Qwen1.5-14B} & Arithmetic & 63.30 & \textbf{71.97} & 71.57 & 69.94 & 66.85 \\
                    & Commensense & 74.43 & \textbf{75.65} & 75.14 & 75.25 & 74.20 \\
                    & Logical & 53.47 & \textbf{55.76} & 54.91 & 55.50 & 54.38 \\
 \midrule
 \multirow{3}{*}{Qwen1.5-1.8B} & Arithmetic & 39.40 & \textbf{39.99} & 37.12 & 31.97 & 28.81 \\
                    & Commonsense & \textbf{57.61} & 55.07 & 57.19 & 55.95 & 55.20 \\
                    & Logical & \textbf{33.00} & 30.72 & 31.00 & 32.37 & 32.05 \\

 \bottomrule
\end{tabular}
}
\caption{Experiment results concerning different tasks. Detailed results are in Appendix~\ref{app:results}. All results are in \%, the best ones are in \textbf{bold}.}
\label{table:results}
\end{table*}

\paragraph{Few-Shot}
Table~\ref{table:btc} on the left side shows comparative performance between traditional few-shot CoT and our "breaking the chain" approach across datasets in commonsense, arithmetic, and logical reasoning tasks. Notably, in arithmetic reasoning, performance on the MultiArith dataset decreases slightly from 99.00\% to 98.33\% with "breaking the chain", while in GSM8K, the decrease is marginal, from 74.60\% to 74.22\%. In commonsense reasoning, "breaking the chain" slightly outperforms the traditional approach on CommonsenseQA (75.18\% vs. 74.94\%), but underperforms on StrategyQA, dropping from 69.13\% to 68.60\%. LogiQA in logical reasoning shows a more notable performance drop from 35.94\% to 33.95\%. These results suggest that while "breaking the chain" generally performs comparably to the few-shot CoT baseline, it does not significantly impact the model's overall performance.

\paragraph{Zero-Shot}
The right side of Table~\ref{table:btc} presents results from our zero-shot probing experiment, comparing the zero-shot CoT baseline with our "break the chain" prompts across 11 datasets within three key tasks: arithmetic reasoning, commonsense reasoning, and logical reasoning. Notably, even when we ablate step-by-step reasoning, ChatGPT maintains competitive performance across various tasks. Moreover, prompting with only "More Tokens" leads to the best performance on several other datasets.

Results for the "Shortcut Reasoning" prompts are detailed in Table~\ref{table:results}, where this approach shows substantial improvements: a 22\% increase in arithmetic tasks, a 9\% boost in commonsense tasks, and an 11\% enhancement in logical reasoning tasks. Performance is consistent on the Mixtral and Qwen platforms, though it varies with the Llama models, underlining the effectiveness of our approach.

In addition, experiments with Qwen models of varying sizes, both under CoT and "break the chain" conditions, are documented. Figure~\ref{fig:model_size} in Appendix~\ref{app:results} illustrates that smaller models exhibit a more pronounced reliance on CoT, especially as the model size decreases, narrowing performance gaps from a 16\% deficit in 72B models to parity in 1.8B models for arithmetic tasks. For logic and commonsense tasks, smaller models transition from underperformance to outperforming larger counterparts, suggesting less capable models benefit more from CoT's structured approach.

These findings question the prevailing assumption that CoT invariably enhances LLM performance. Our results indicate that specific prompts, even without detailed reasoning, can yield comparable or superior outcomes. However, the effectiveness of "break the chain" prompts varies, pointing to a nuanced interplay between prompt nature and LLM performance that merits further investigation.

We observe that CoT is particularly adept at tackling questions decomposable into sub-issues that are solvable in brief sentences. Challenges arise when generated responses become excessively lengthy, leading to potential task misalignment and illogical outputs, or when they exceed the maximum length constraints set in the code, inhibiting the completion of reasoning sequences.

\begin{table*}
\centering
\scalebox{0.75}{
\begin{tabular}{c|c|c|c|c|c|c}
 \toprule
 Dataset  & Question Type & Base & Quick Conclude & Shortcut Reasoning & Effective Shortcut & Innovative Shortcut  \\ 
 \midrule
 \multirow{3}{*}{ShortcutQA} & Analytical Reasoning & 26.79 & \textbf{33.93} & 21.43 & 19.64 & 30.36\\ 
   & Verbal Reasoning & 22.73 & \textbf{25.00} & 22.73 & 23.86 & 21.59 \\ 
  & Mathematical Reasoning & 25.00 & 30.95 & 29.76 & 26.19 & \textbf{32.14} \\
 \bottomrule

\end{tabular}
}
\caption{Performance comparison across tasks within ShortcutQA.}
\label{table:shorcutQA_res}
\end{table*}

\paragraph{ShortcutQA}
Table~\ref{table:shorcutQA_res} presents a comparative analysis of performance across various task types within the ShortcutQA dataset. Compared to benchmarks utilized elsewhere in this study, ShortcutQA poses a greater challenge, making it an ideal testing ground for advancing model capabilities.

In mathematical reasoning tasks, all "break the chain" prompts outperform the established baselines. The "Innovative Shortcut" prompt is particularly effective, achieving a significant relative improvement of 28.56\% over the baseline. "Quick Conclude" also shows substantial gains, with a relative increase of 23.8\% compared to the baseline.

For analytical and verbal reasoning tasks, "Quick Conclude" registers the highest improvements, with increases of 26.65\% and 9.99\%, respectively, over the baseline. "Innovative Shortcut" also posts notable gains in analytical tasks, while "Effective Shortcut" sees considerable enhancements in verbal tasks.

Overall, "Innovative Shortcut" and "Quick Conclude" are standout performers on the ShortcutQA dataset, underscoring the potency of our "break the chain" strategy. This dataset not only challenges current LLMs but also sets a benchmark for future enhancements, providing a robust platform for testing and refining next-generation models.

\section{Discussion}

\subsection{Reasoning with Token Limits}
\begin{figure}[h]
    \centering
    \includegraphics[width=0.5\textwidth]{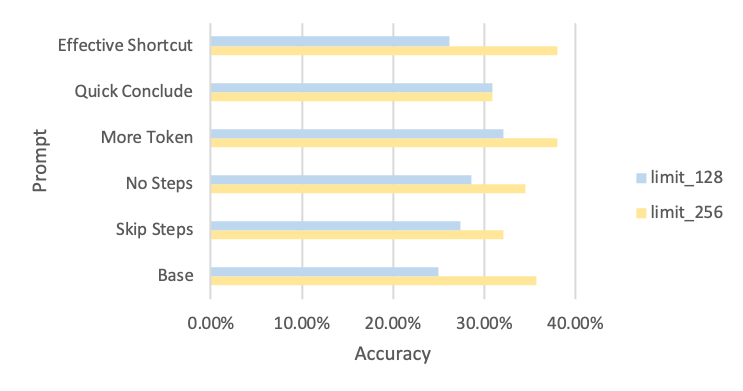}
    \caption{Performance comparison of different token limits on the mathematical reasoning task from ShortcutQA.}
    \label{fig:token_limit}
\end{figure}

We investigated the impact of token limits on model performance by experimenting with different constraints (128, 256 tokens) during the response generation phase. Figure~\ref{fig:token_limit} illustrates how varying token limits affect outcomes on the mathematical reasoning task within ShortcutQA using different prompts. We observed that as the token limit increases, so does performance across all prompts, indicating that constraints on output length significantly influence the inference process and thus the results. Notably, even at the minimum limit of 128 tokens, all prompts exceed the baseline performance, suggesting that our "break the chain" approach is not only efficient but also effective in conserving computational resources while maintaining or improving task performance.

\subsection{Theoretical Analysis}
We have developed a qualitative model to formalize the performance dynamics of Chain-of-Thought (CoT) reasoning and to elucidate the effectiveness of the "Break the Chain" approach.

In our framework, each CoT step is divided into two phases: analysis and reasoning. The accuracy of the analysis at step $t$ is denoted as $P(a_t)$, and the subsequent reasoning based on this analysis is denoted as $P(r_t)$. Therefore, the total accuracy of a CoT sequence depends on the combined accuracy of these phases across all steps, mathematically expressed as:

\begin{equation}
P(\text{CorrectReasoning}_{\text{CoT}}) = \prod_{t=1}^{T} P(a_t) P(r_t),
\label{eq:1}
\end{equation}

where $T$ is the total number of steps in the CoT reasoning chain. To evaluate the efficacy of different prompting strategies, we define $P(\text{CorrectReasoning}_{p})$ as the probability of achieving correct reasoning for a given prompt $p$. A prompt is considered more effective than the traditional CoT approach if $P(\text{CorrectReasoning}_{p})$ surpasses $P(\text{CorrectReasoning}_{\text{CoT}})$.

In cases where no explicit analysis or reasoning phase is involved, and both are integrated by LLMs in each step, Equation~\ref{eq:1} simplifies to:

\begin{equation}
P(\text{CorrectReasoning}_{\text{CoT}}) = \prod_{t=1}^{T} P(i_t),
\label{eq:2}
\end{equation}

where $i_t$ signifies the probability of obtaining the correct result in a single, consolidated inference step.

\begin{figure}[h]
    \centering
    \includegraphics[width=0.5\textwidth]{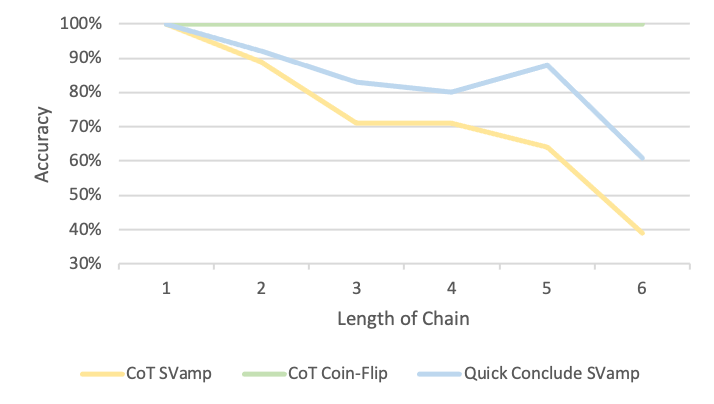}
    \caption{Relationship between CoT Chain Length and Accuracy.}
    \label{fig:impact on length of chain}
\end{figure}

Our experimental results corroborate the theoretical predictions, as illustrated in Figure~\ref{fig:impact on length of chain}. We observe that CoT accuracy generally declines as chain length increases. Notably, in scenarios like Coin Flip where $P(i_t)$ approaches 1, accuracy remains stable regardless of chain length. Conversely, in tasks like SVamp where $P(i_t)$ is lower, a decrease in accuracy is noted as the chain lengthens. When comparing "Quick Conclude" on SVamp against baseline accuracies, the relative CoT accuracy diminishes with increasing chain length, aligning precisely with our model. Detailed methodologies for these experiments are available in Appendix~\ref{app:methods}.



\section{Conclusion}
This study critically evaluates Chain-of-Thought (CoT) reasoning in language models, highlighting limitations such as high token consumption and limited applicability. Our "break the chain" strategies integrate human-like heuristics and shortcuts, enhancing efficiency without compromising performance across various models. The introduction of the ShortcutQA dataset further advances AI reasoning evaluation by focusing on heuristic tasks, providing a robust benchmark that challenges traditional methods. Our findings suggest that adopting more intuitive, efficient reasoning approaches could significantly improve the problem-solving capabilities of AI systems in real-world applications.

\bibliography{custom}

\appendix

\section{Zero-shot prompts for "break the chain"}
\label{app:prompts}
The abbreviations of probing prompts and shortcut prompts are shown in the table~\ref{table:detailed_prompt}.

\begin{table*}
\centering
\scalebox{0.8}{
\begin{tabular}{c|c|c} 
 \toprule
 Prompt Type & Abbreviations & Full Prompts  \\ 
 \midrule

 \multirow{3}{*}{Probing Prompts} & Skip Steps & Let's skip as much as possible.  \\ 
  \cmidrule{2-3} 
  & No Steps & Let's don't think step by step.  \\
  \cmidrule{2-3} 
  & More Token & Let's think as much as possible. \\

 \midrule
 \multirow{4}{*}{Shortcut Prompts} & Quick Conclude & Let's quickly conclude the answer without showing step-by-step reasoning. \\ 
  \cmidrule{2-3}
  & Shortcut Reasoning & Let's quickly conclude the answer with shortcut reasoning. \\
 \cmidrule{2-3}
  & Effective Shortcut & Rapidly evaluate and use the most effective reasoning shortcut to answer the question. \\
  \cmidrule{2-3}
   & Innovative Shortcut & Think outside the box and quickly identify an innovative shortcut to solve this problem. \\

 \bottomrule
\end{tabular}
}
\caption{The relationship between a prompt abbreviation and its full prompt.}
\label{table:detailed_prompt}
\end{table*}

\section{Pipeline Details}
\label{app:pipeline}
Figure~{fig:method} shows the pipeline of experiments.
\begin{figure*}
  \centering
  \includegraphics[width=\textwidth]{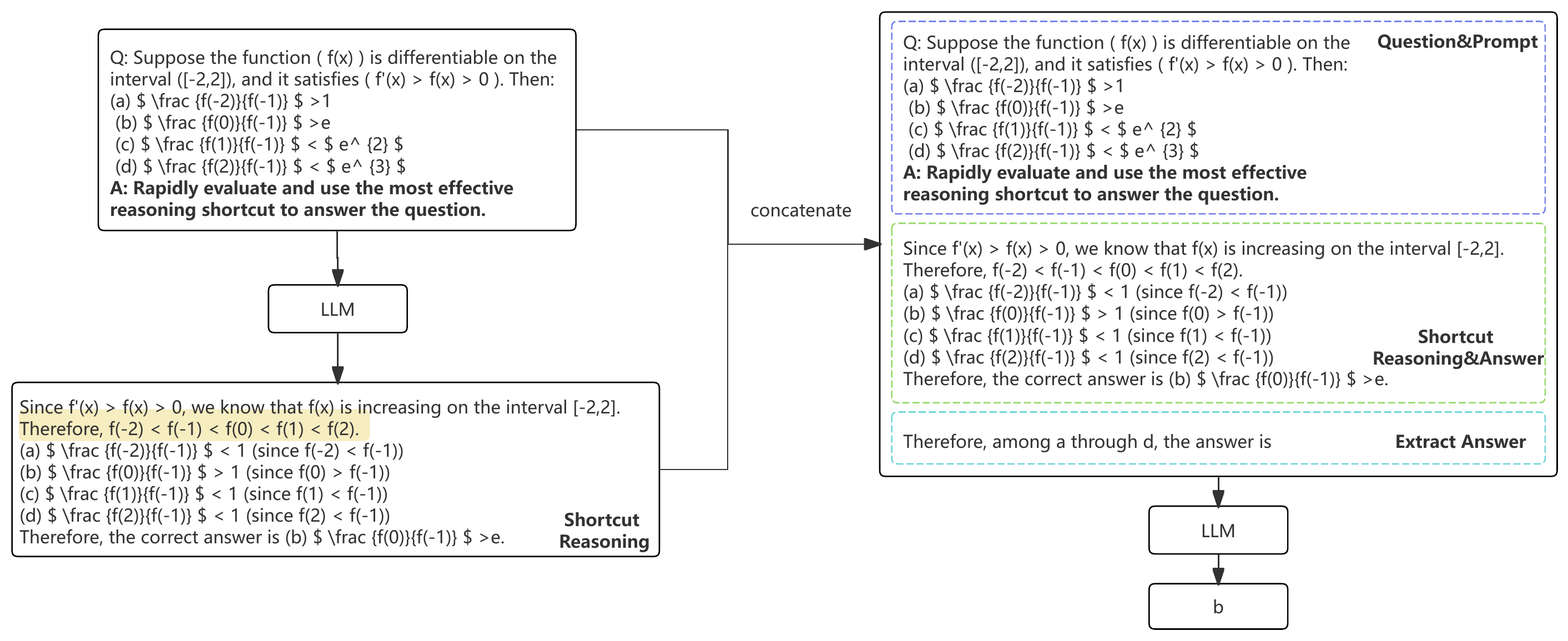}
  \caption{Our evaluation pipeline.}
  \label{fig:method}
\end{figure*}

\section{Experiment Results}
\label{app:results}

Table~\ref{table:performance_comparison} is the original experiment results of diverse model structures.

\begin{figure}[h]
    \centering
    \includegraphics[width=0.5\textwidth]{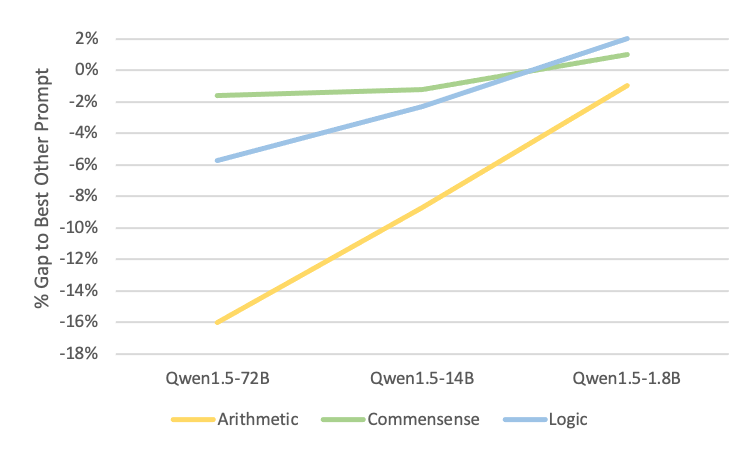}
    \caption{The Impact of Model Size on CoT's Relative Outperformance over Other Prompts across Datasets}
    \label{fig:model_size}
\end{figure}

Figure~\ref{fig:model_size} shows that as model size decreases, CoT's relative performance advantage over other prompts increases across all tasks.

\section{Detailed Methods}
\label{app:methods}
In this section, we introduce our detailed methods for our experiments.
For our experiment in discussion, we generally used GPT4 to evaluate the logs, and caculate the accuracy of different lengths. First, We used GPT4 to check the logs of CoT to calculate the length of chain in each question on SVamp and Coin Flip. Second, we calculated the accuracy at different length of chain. Third, to exclude the disturbance of various difficulty distributions within each group, we calculated the accuracy with promptQC in each group of data as baseline without CoT on SVamp.

\begin{table*}
\centering
\scalebox{0.7}{
\begin{tabular}{c|l|l|l|l|l|l|l}
 \toprule
 model  & Task  & Dataset & Base & Quick Conclude & Shortcut Reasoning & Effective Shortcut & Innovative Shortcut \\ 
 \midrule
 \multirow{7}{*}{ChatGPT} & \multirow{6}{*}{Arithmetic} & SingleEq & 86.61 & 91.14 & 91.73 & \textbf{92.32} & 77.36 \\
                           &                    & AddSub & 83.80 & \textbf{90.89} & 86.33 & 89.62 & 73.67 \\
                           &                    & AQUA-RAT & 35.43 & 51.97 & 52.76 & \textbf{53.94} & 52.36 \\
                           &                    & MultiArith & 83.33 & 91.00 & 94.67 & \textbf{94.83} & 89.83 \\
                           &                    & GSM8K & 32.68 & 56.86 & \textbf{71.57} & 67.25 & 58.00 \\
                           &                    & SVAMP & 71.70 & 81.50 & 83.60 & \textbf{85.50} & 76.80 \\
\cmidrule{2-8}             
                           & \multirow{2}{*}{Commensense} & CommonsenseQA & 70.52 & 77.89 & \textbf{78.95} & 76.82 & 72.89 \\
                           &                    & StrategyQA & 64.37 & \textbf{68.47} & 68.34 & 67.90 & 62.14 \\
\cmidrule{2-8}             
                           & \multirow{2}{*}{Logic} & LogiQA & 40.09 & 41.32 & \textbf{43.32} & 42.70 & 41.01 \\
                           &                    & ReClor & 52.40 & \textbf{52.80} & 51.60 & 52.00 & 51.40 \\
                           &                    & Date Understanding & 63.41 & 65.85 & 77.78 & 75.61 & \textbf{78.32} \\
 \midrule
 \multirow{7}{*}{Llama-70B} & \multirow{6}{*}{Arithmetic} & SingleEq & 67.91 & 55.91 & \textbf{81.10} & 49.80 & 40.16 \\
                           &                    & AddSub & 69.87 & 40.51 & \textbf{80.25} & 53.92 & 32.66 \\
                           &                    & AQUA-RAT & 61.02 & 52.36 & \textbf{62.20} & 57.87 & 50.79 \\
                           &                    & MultiArith & 79.83 & 71.00 & \textbf{93.33} & 73.67 & 60.33 \\
                           &                    & GSM8K & 80.97 & 81.05 & \textbf{85.14} & 78.17 & 69.52 \\
                           &                    & SVAMP & 75.20 & 72.90 & \textbf{87.50} & 66.80 & 52.30 \\
\cmidrule{2-8}             
                           & \multirow{2}{*}{Commensense} & CommonsenseQA & 79.03 & \textbf{81.49} & 77.31 & 69.21 & 74.37 \\
                           &                    & StrategyQA & 56.11 & 64.50 & 43.84 & \textbf{65.37} & 60.17 \\
\cmidrule{2-8}             
                           & \multirow{2}{*}{Logic} & LogiQA & 57.60 & 57.30 & 57.45 & \textbf{58.83} & 56.07 \\
                           &                    & ReClor & \textbf{71.80} & 69.40 & 68.40 & 69.80 & 70.20 \\
                           &                    & Date Understanding & \textbf{84.82} & 72.09 & 79.95 & 72.90 & 65.58 \\
 \midrule
 \multirow{7}{*}{Mixtral-8x7B} & \multirow{6}{*}{Arithmetic} & SingleEq & 87.40 & \textbf{88.58} & 87.01 & 83.46 & 70.87 \\
                           &                    & AddSub & 85.82 & \textbf{86.84} & 84.81 & 83.54 & 72.15 \\
                           &                    & AQUA-RAT & 37.40 & 41.34 & \textbf{42.13} & 39.76 & 32.68 \\
                           &                    & MultiArith & \textbf{87.50} & 87.33 & 85.83 & 78.17 & 62.17 \\
                           &                    & GSM8K & 48.90 & \textbf{55.80} & 54.81 & 49.20 & 39.12 \\
                           &                    & SVAMP & 77.80 & \textbf{79.40} & 75.60 & 78.50 & 62.80 \\
\cmidrule{2-8}             
                           & \multirow{2}{*}{Commensense} & CommonsenseQA & 71.63 & 72.40 & \textbf{72.73} & 71.17 & 65.85 \\
                           &                    & StrategyQA  & 58.43 & 66.33 & 65.72 & \textbf{67.82} & 55.76 \\
\cmidrule{2-8}             
                           & \multirow{2}{*}{Logic} & LogiQA & 42.70 & 45.01 & 38.56 & 40.86 & \textbf{45.78} \\
                           &                    & ReClor & 47.40 & 50.60 & \textbf{51.80} & 48.60 & 47.60 \\
                           &                    & Date Understanding & 66.40 & 67.48 & 63.96 & \textbf{68.02} & 59.08 \\
\midrule
 \multirow{7}{*}{Qwen1.5-72B} & \multirow{6}{*}{Arithmetic} & SingleEq & 80.71 & 87.80 & \textbf{88.78} & 88.58 & 86.61 \\
                           &                    & AddSub & 84.56 & 84.81 & 86.33 & \textbf{88.61} & 86.84 \\
                           &                    & AQUA-RAT & 37.80 & 47.24 & \textbf{48.43} & 46.46 & 35.43 \\
                           &                    & MultiArith & 81.33 & \textbf{96.00} & 95.33 & \textbf{96.00} & 93.67 \\
                           &                    & GSM8K & 28.96 & \textbf{54.06} & 48.98 & 45.26 & 42.30 \\
                           &                    & SVAMP & 78.30 & \textbf{86.10} & 85.20 & 82.20 & 80.10 \\
\cmidrule{2-8}             
                           & \multirow{2}{*}{Commensense} & CommonsenseQA & 81.98 & 83.54 & 81.98 & 83.37 & \textbf{83.7} \\
                           &                    & StrategyQA  & 76.24 & 76.16 & 76.77 & \textbf{77.34} & 75.85 \\
\cmidrule{2-8}             
                           & \multirow{2}{*}{Logic} & LogiQA & 46.54 & 50.08 & \textbf{51.15} & 50.54 & 47.00 \\
                           &                    & ReClor & 61.60 & \textbf{66.20} & 64.00 & 65.80 & 64.40 \\
                           &                    & Date Understanding & 72.36 & 73.98 & \textbf{75.61} & 74.53 & 73.98 \\
 \bottomrule
\end{tabular}
}
\caption{Comparison of Various Open-Source Large Models' Performance with Different Prompts Across Multiple Datasets.}
\label{table:performance_comparison}
\end{table*}


\end{document}